\documentclass[conference]{IEEEtran}
\usepackage{gensymb}
\usepackage{amsmath,amssymb,amsfonts}
\usepackage{algorithmic}
\usepackage{graphicx}
\usepackage{hyperref, color}
\usepackage{textcomp}
\usepackage{multicol, multirow}
\usepackage{comment}
\usepackage{xcolor}
\usepackage{subcaption}
\usepackage{array}
\newcolumntype{M}[1]{>{\centering\arraybackslash}m{#1}}
\def\BibTeX{{\rm B\kern-.05em{\sc i\kern-.025em b}\kern-.08em
    T\kern-.1667em\lower.7ex\hbox{E}\kern-.125emX}}
\begin{document}

\title{OSSR-PID: \textbf{O}ne-\textbf{S}hot \textbf{S}ymbol \textbf{R}ecognition in \textbf{P}\&\textbf{ID} Sheets using Path Sampling and GCN
}

\author{\IEEEauthorblockN{Shubham Paliwal, Monika Sharma and Lovekesh Vig}
\IEEEauthorblockA{\textit{TCS Research, India} \\
Email: {\{shubham.p3, monika.sharma1, lovekesh.vig\}@tcs.com}}
}

\maketitle

\begin{abstract}
  In this paper, we focus on recognition of line-drawn symbols in engineering drawings with only one prototypical example per symbol available for training. In particular, Piping and Instrumentation Diagrams (P\&ID) are ubiquitous in several manufacturing, oil and gas enterprises for representing engineering schematics and equipment layout. There is an urgent need to extract and digitize information from P\&IDs without the cost of annotating a varying set of symbols for each new use case. A robust one-shot learning approach for symbol recognition i.e., localization followed by classification, would therefore go a long way towards this goal. 
  Our method works by sampling pixels sequentially along the different contour boundaries in the image. These sampled points form paths which are used in the prototypical line diagram to construct a graph that captures the structure of the contours. Subsequently, the prototypical graphs are fed into a Dynamic Graph Convolutional Neural Network (DGCNN) which is trained to classify graphs into one of the given symbol classes. Further, we append embeddings from a Resnet-34 network which is trained on symbol images containing sampled points to make the classification network more robust. Since, many symbols in P\&ID are structurally very similar to each other, we utilize Arcface loss during DGCNN training which helps in maximizing symbol class separability by producing highly discriminative embeddings. During inference time, a similar line based sampling procedure is adopted for generating sampled points across P\&ID image. The images consist of components attached on the pipeline (straight line). The sampled points segregated around the symbol regions are used for the classification task.  
  The proposed pipeline, named \textbf{OSSR-PID}, is fast and gives outstanding performance for recognition of symbols on a synthetic dataset of $100$ P\&ID diagrams. We also compare our method against prior-work that uses full supervision (not one-shot) on a real-world private dataset of $12$ P\&ID sheets and obtain comparable/superior results. Remarkably, it is able to achieve such excellent performance using only one prototypical example per symbol.
\end{abstract}


\section{Introduction}
Piping and instrumentation diagrams (P\&ID)~\cite{pid_1} are a standardized format for depicting detailed schematics about material flow, equipment components, and control devices in oil and gas, chemical manufacturing and underground utility companies. A P\&ID is based on process flow diagrams which illustrate the piping schematics with instruments using line and graphical symbols, along with appropriate annotations. Presently, millions of such legacy P\&IDs are stored in an unstructured image format and the data trapped in these documents is required for inventory management, detailed design, procurement, and construction of a plant. Manually extracting information from such P\&ID sheets is a very tedious, time-consuming and error-prone process as it requires that users interpret the meaning of different components such as pipelines, inlets, outlets and graphic symbols based on the annotations, and geometric and topological relationships of visual elements~\cite{pid_2}. To mitigate these issues, there is an urgent need to automate the process of digitizing P\&ID sheets to facilitate fast, robust and quick extraction of information.

The major components of P\&ID include symbols with field-specific meanings, pipelines representing connections between symbols, and textual attributes~\cite{pid_3}. There exists very limited body of work on digitization of engineering drawing documents in the literature ~\cite{work1, work2, auto_pid1, arindam, gaurav}. Early attempts used traditional image recognition techniques utilizing geometrical features of objects such as edge detection~\cite{edge}, hough transform~\cite{hough} and morphological operations~\cite{morpho}. Authors in ~\cite{case-based-ed} proposed an approach where the system tries to learn and store the graphical representation of the symbol presented by the user. Later, the system uses the acquired knowledge about the symbol for recognition. These earlier systems based on standard vision techniques were not powerful enough to address the challenges arising due to noisy images, variations in text fonts, resolutions and minute visual differences among various symbols.

\begin{figure*}[h]
\centering
\includegraphics[width=0.8\textwidth]{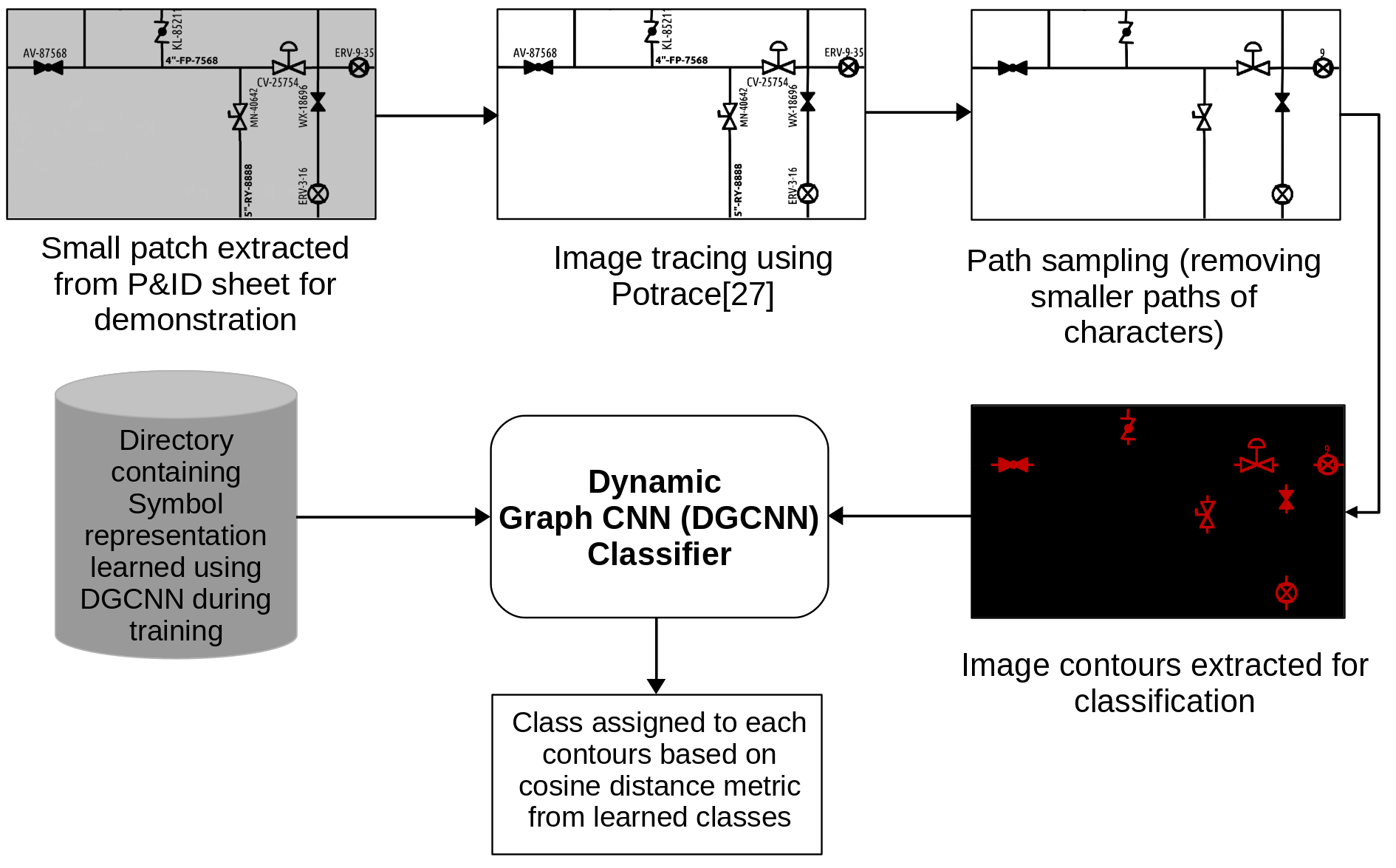}
\caption{Flowchart showing different modules of OSSR-PID proposed for Symbol Recognition in P\&ID sheets.}
\label{fig:flowchart}
\end{figure*}

Recently deep learning techniques~\cite{deep1} have yielded significant improvements in accuracy for information extraction from P\&IDs. Recent work by Rahul et. al.~\cite{auto_pid2} utilize a combination of traditional vision techniques and state-of-the-art deep learning models to identify and isolate pipeline codes, pipelines, inlets and outlets, and for detecting symbols. The authors proposed the use of Fully Convolutional Networks (FCN) based segmentation for detection of symbols to address low inter-class variability and noisy textual elements present inside symbols. However, this requires significant volumes of annotated symbols for training the network and fails for rarely occurring symbols for which training data is sparse. Additionally, in practice, the symbol-set keeps changing and ideally, the system should allow for introduction of new symbols without manual annotation or complete re-training. In effect, given just one clean symbol image, we wish to detect all instances of the symbol in noisy and crowded P\&ID diagrams.

To address these challenges, we propose to exploit the recent advances in few-shot/one-shot learning ~\cite{liu2019few, SnellSZ17, SatorrasE18, Dong2019OneShotNA, Vinyals2016MatchingNF}. Recent attempts at one-shot learning broadly adopt two approaches: 1) metric learning~\cite{Wang2019MultiSimilarityLW, metric, Vinyals2016MatchingNF} which involves learning an embedding space to separate different classes or 2) a meta-training phase~\cite{meta-learning, meta-sgd} which involves training a meta learner on a collection of tasks from the same task distribution to discover a task-independent network initialization. However, such techniques require a significant number of training tasks for meta-training prior to being employed for one-shot recognition. We wish to avoid the meta-training step entirely, and to directly perform one-shot symbol recognition. Hence, we propose a method for recognition of symbols with just one training sample per class, which is represented as a graph with points sampled along the boundaries of different entities present in the P\&ID and subsequently, utilize Dynamic Graph Convolutional networks~\cite{dgcnn} for symbol classification. In particular, we make the following contributions in this paper:
\begin{itemize}
    \item We propose a novel, fast and efficient one-shot symbol recognition method, \textit{OSSR-PID}, for P\&ID sheets utilizing only one prototypical example image per symbol class.
    \item We propose a method to sample pixels along the boundaries of different entities, which are present in P\&ID image and construct a graph which captures the connectivity between different entities (instruments / symbols).
    \item We utilize Dynamic Graph Convolutional Neural Networks (DGCNNs)~\cite{dgcnn} for symbol graph classification. We also augment the network with visual embeddings from Resnet-34 model trained on symbol-images of sampled points to make it more robust.
    \item We utilize Arcface loss~\cite{deng2019arcface} for training of DGCNN network for the classification task as it optimizes feature embeddings to enforce higher similarity for intra-class symbols and diversity for inter-class symbols.
    \item We evaluate OSSR-PID for symbol recognition on a synthetic dataset called Dataset-P\&ID~\cite{paliwal_jain_sharma_vig_2021}~\footnote{
\label{data:link}
Dataset-P\&ID:
\url{https://drive.google.com/drive/u/1/folders/1gMm_YKBZtXB3qUKUpI-LF1HE_MgzwfeR}} containing $100$ P\&ID sheets in the test-set.
    \item We also compare OSSR-PID against prior work by \cite{auto_pid2} on a private dataset of $12$ real-world P\&IDs and achieve remarkable results.
\end{itemize}

Rest of the paper is organized as follows: Section~\ref{sec:problem-formulation} gives an overview of the problem that we solve in this paper. Details of dynamic graph CNN, its definition and edge convolutions are given in Section~\ref{sec:dynamic-graph}. Subsequently, we describe the approach adopted for one shot symbol recognition in Section~\ref{sec:methodology} by explaining each step such as image tracing, path sampling, symbol-region segregation and graph convolutional network based symbol classification in Section~\ref{subsec:image-tracing}~\ref{subsec:path-sampling},~\ref{subsec:symbol-reg-seg} and ~\ref{subsec:gcn-identification} respectively. Section~\ref{sec:experimental-results} presents details of the dataset, results of the experiments conducted followed by discussions on them. 
Finally, we conclude the paper in Section~\ref{sec:conclusion}.

\begin{figure*}
\centering
\begin{subfigure}[b]{0.45\textwidth}
\includegraphics[width=\textwidth]{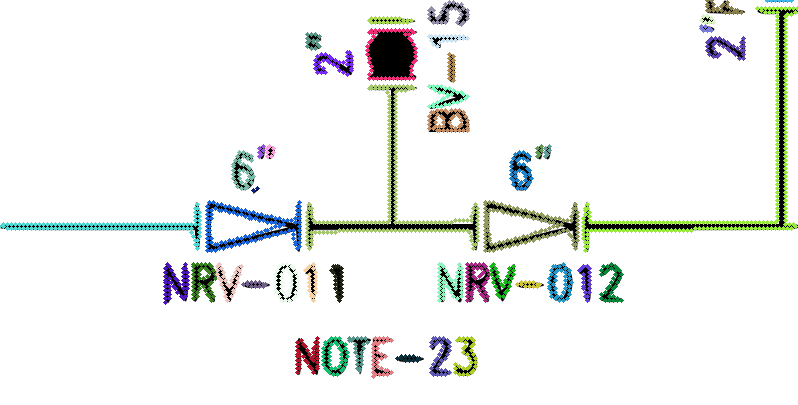}
\caption{Individual paths obtained are plotted in different colors.}
\label{fig:sam_paths_1loop}
\end{subfigure}
\begin{subfigure}[b]{0.45\textwidth}
\includegraphics[width=\textwidth]{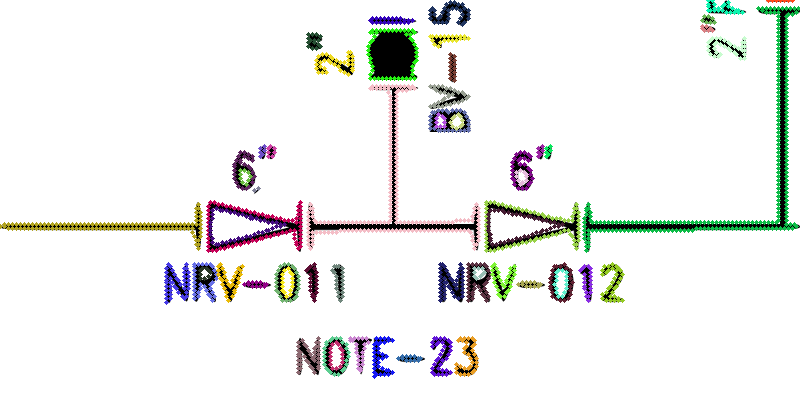}
\caption{Different loops from each path, plotted in different colors.}
\label{fig:sam_paths}
\end{subfigure}
\caption{Figures illustrating image-tracing output (a) Different paths obtained represented with different colors. (b) Different sequential loops (sampled points) obtained from each path covering the outline of image entities. Please note that in each path, there are multiple sequential loops over every image entity. As evident by the alphabet 'O' in "NOTE-23" in (a), entire alphabet is created using single path (represented in green color). A single sequential loop cannot cover both interior and exterior of character 'O'. Thus, there are two separate sequential loops on the outline as shown with different colors in (b).}
\end{figure*}

\section{Problem Formulation}
\label{sec:problem-formulation}
This paper proposes a method to process scanned P\&ID sheets and detect the different symbols (components) present in each sheet. Additionally, we wish to find connections between components and identify different pipelines. Identification of pipeline involves sampling points along the periphery of the dark regions(printed) of P\&ID sheets. Further, the regions are segmented from the points which form straight lines (i.e. pipes) to get the probable symbol regions. Subsequently, the points present in these regions are classified using Dynamic Graph Convolutional Neural Networks (DGCNNs)~\cite{dgcnn} as belonging to one of the given symbol classes. 

\section{Dynamic Graph Convolutional Neural Network}
\label{sec:dynamic-graph}
DGCNN~\cite{dgcnn} takes given set of input data points to create a local graph in which each data point is connected to its $n$ nearest data points. The edges formed between the nearest neighbors are used for convolutional operations similar to GNNs~\cite{wu2019comprehensive}. As the name suggests, the graph created initially from the data is dynamic. Every layer of the network uses a k-NN algorithm to obtain a new set of $n$ neighbors based upon the data point's embedding at the respective layer. Readers are referred to ~\cite{qi2016pointnet}~\cite{dgcnn} for more details.

\subsection{Graph definition}
\label{subsec:graph-def}
Consider a directed graph $G = (V, E)$, where $V=\{1,...,n\}$ represents vertices of the graph and $E \subseteq V \times V$ represents edges that connect each vertex to its k-nearest neighbours. Each vertex $v$ has an $F$-dimensional feature representation. Next, edge features are defined as $e_{ij} = h_{\Theta}(v_{i}, v_{j})$, where $h_{\Theta}: R^{F} \times R^{F} \to R^{F^{'}} $ are non-linear operations with set of learnable parameters $\Theta$. The constructed graph also consists of an additional edge with itself (i.e., self-loop) so that every vertex is pointing to itself.

\subsection{Edge Convolution}
\label{subsec:edge-conv}
\begin{figure*}[h]
\centering
\begin{subfigure}[b]{0.74\textwidth}
\includegraphics[width=\textwidth]{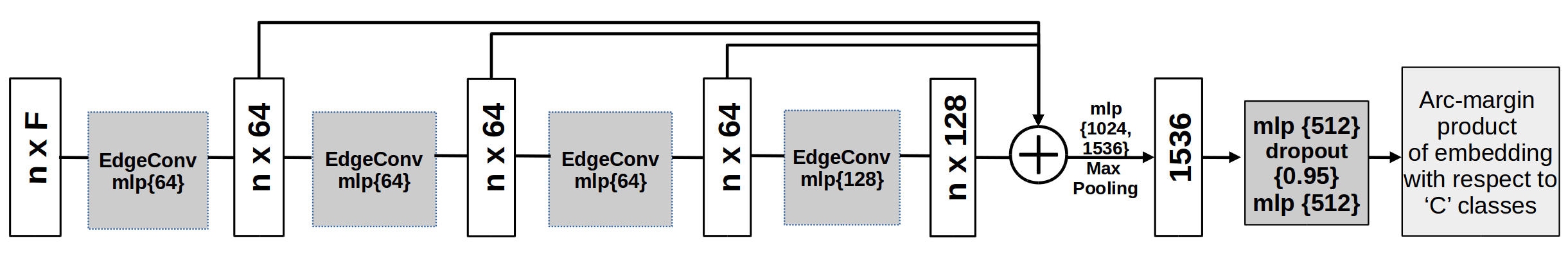}
\caption{DGCNN Architecture}
\end{subfigure}
\begin{subfigure}[b]{0.24\textwidth}
\includegraphics[width=\textwidth]{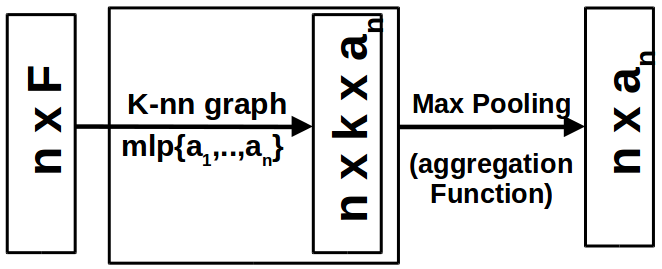}
\caption{EdgeConv block}
\end{subfigure}
\caption{\textbf{(a)} DGCNN~\cite{dgcnn} architecture, used for classification of sampled points into one of the symbol classes, takes $F=9$-dimensional $n$=1024 points as input. The network comprises of a set of $3$ EdgeConv layers whose outputs are aggregated globally to get a global descriptor, which is later used to obtain classification scores for $C$ classes. \textbf{(b)} EdgeConv block: The EdgeConv layer takes input tensor of shape $n \times F\textsubscript{l}$ and finds $k$ nearest neighbors based upon distance between embeddings. Subsequently, edge features are computed on these $k$ edges by applying multi-layer perceptron on $\{a\textsubscript{1},...,a\textsubscript{n}\}$ and the output tensor of shape $n \times k \times a\textsubscript{n}$  is max pooled over $k$ neighbors, resulting in final output tensor of shape of $n \times a\textsubscript{n}$.}
\label{fig:network_arch}
\end{figure*}

At each layer $l$, the connectivity of every vertex is determined using a k-NN algorithm based on the feature embeddings of the vertices. For every pair of connected vertices $v_{i}$ and $v_{j}$, features of edges $e_{ij}$ are computed as explained in Section~\ref{subsec:graph-def}. Subsequently, the edge features computed across the neighboring edges of a vertex are aggregated via a $max$ operator to update its embedding feature. Thus, the output obtained after the edge convolution is given by
\begin{equation}
  v_{i}^{'} = \max_{j:(i,j) \in \varepsilon} h_{\Theta}(v_{i}, v_{j})
\end{equation}
However, to capture the local information as well as the global information, DGCNN defines $h_{\Theta}(v_{i}, v_{j}) = h_{\Theta}(v_{i}, v_{j}-v_{i})$, where $v_{i}$ defines the global information of the feature point and $v_{j}-v_{i}$ defines the local relative information of the feature point. Mathematically, this is represented as follows:
\begin{equation}
  v_{i}^{'} = \max_{j:(i,j) \in \varepsilon} 
  ReLU({\theta}_m . (v_{j}-v_{i}) + {\phi}_m . v_{i})
\end{equation}
where, $\varepsilon$ represents the set of neighbors to vertex $v_{i}$. Figure~\ref{fig:network_arch} represents the Dynamic Graph CNN architecture used for symbol classification.

\begin{figure*}[h]
\centering
\includegraphics[width=\textwidth]{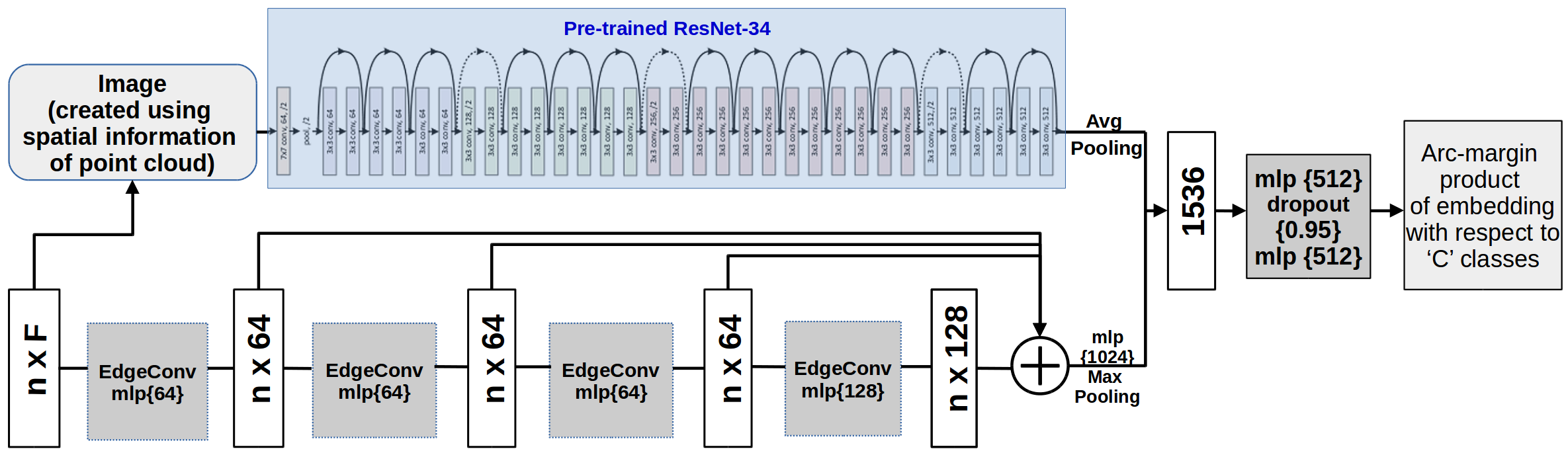}
\caption{\textbf{DGCNN+Resnet-34+ArcFace Loss}: Pre-trained Resnet-34 is used for extracting visual features from images($224 \times 224$) created from spatial information of point cloud. Subsequently, these Resnet-34 embeddings are aggregated with outputs of EdgeConv layers of DGCNN to obtain global descriptors, which are finally utilized to generate the embeddings. These embeddings are trained after introducing arc-margin with respect to their corresponding symbol classes.}
\label{fig:dgcnn_resnet}
\end{figure*}

\section{Methodology}
\label{sec:methodology}
In this section, we describe the proposed pipeline for symbol recognition, as shown in Figure~\ref{fig:flowchart}, from P\&ID sheets. Section~\ref{subsec:image-tracing} describes how our image tracing technique produces smooth contours using a series of bezier~\cite{cite-potrace} curves and lines. This is followed by a description of the path sampling technique in Section~\ref{subsec:path-sampling}. A description of how the individual connecting pipes are segregated and potential separate regions are created for symbol identification is provided in Section~\ref{subsec:symbol-reg-seg}. Finally, details of the DGCNN model for one-shot symbol localization and classification of separated contours is provided in Section~\ref{subsec:gcn-identification} and ~\ref{subsec:inference}.

\subsection{Image tracing}
\label{subsec:image-tracing}

The black and white pixels in a binarized image give rise to several contours with dark regions forming separate contours. In image tracing, we process the image to produce a vector outline of these contours by approximating them via a set of algebraic equations like bezier curves~\cite{cite-potrace}. The spline curves generalize the contours by creating smoother outlines. We use adaptive thresholding~\cite{gonzalez2004digital} on the input image with a window size of $13 \times 13$ to efficiently segregate foreground (black pixels) and background (white pixels) regions. The contours for dark regions are obtained by a series of bezier curves and straight lines using the Potrace algorithm\cite{cite-potrace}. The output equations obtained are categorized as paths or loops. Contours which are connected in the diagram, are traced by sequential bezier curves which are connected end-to-end, forming a path. Each path is composed of one or more loops such that their ends are connected. Figure~\ref{fig:sam_paths_1loop}, shows different paths which are created to trace the dark contours using image tracing. 
 
 \subsection{Path sampling}
\label{subsec:path-sampling}
After image tracing, we correct the image for rotation and scaling variations by applying a set of algebraic operations. The images are resized to a fixed width of $4000 px$ while maintaining the aspect ratio. The paths generated from the Potrace algorithm consist of end-to-end bezier curves along the outlines of contours present in the image. We use these curves to sample a set of points which are continuous and are at fixed distance intervals. Since, the points are generated on the periphery of the contour regions, the close adjacent points (across the edges) are merged together. The set of points along each path formed on the outline of the same contour will have strong correlations w.r.t. slope and distance. The regions where the above two parameters vary, are marked as junction regions. Note that the potrace algorithm generates paths which contain loops. Each contour is guaranteed to be covered by the different loops present in the path. We create a unique graph for each path from the obtained set of points. The critical junctions of these graphs are obtained by identifying the points from where a new branch of points emerges or more than one branch of points merge together. 

\begin{figure*}[h]
\centering
\includegraphics[width=0.7\textwidth]{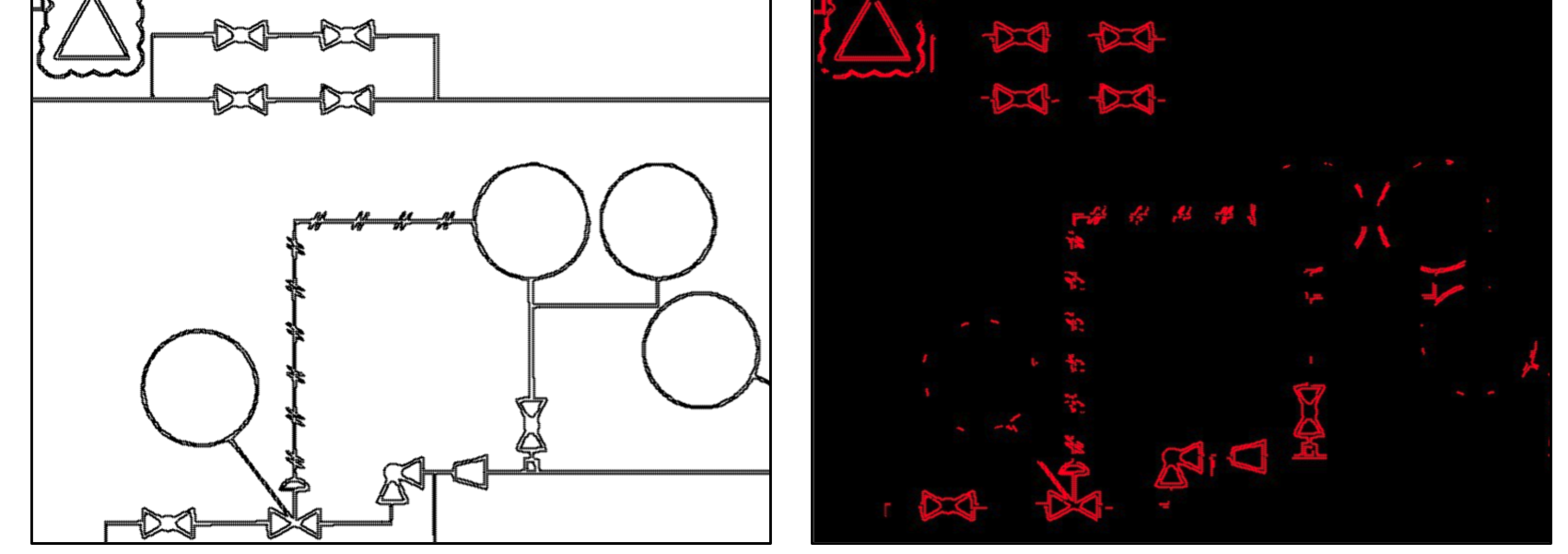}
\caption{\textbf{(Left)} Image patch from P\&ID showing contours. \textbf{(Right)} Extracted symbol regions(red).}
\label{fig:region_sampling}
\end{figure*}

\subsection{Symbol Region Segregation}
\label{subsec:symbol-reg-seg}
The paths obtained using path sampling include pipelines and symbols present in P\&ID. The regions containing symbols are required to be segmented out for classification using Graph Convolutional Network. To achieve this, the obtained individual branches are processed to remove pipes (lines) from the P\&ID image as follows: 
\begin{itemize}
\item The slope of each point $p$ in path is computed by using the average of two neighbors $p_{i-1}$ and $p_{i+1}$ as given by Equation.~\ref{eq:slope_equation}
\begin{equation}
 \hspace{-5mm} \small{slope_i =  
  Avg(atan2(p_{i-1},p), atan2(p,p_{i+1}),atan2(p_{i-1},p_{i+1}))}
\label{eq:slope_equation}
\end{equation}

\item The points along the paths are traversed and for every point, a window of length $1$ and height $t$ in the orthogonal direction to the point's slope, is checked for the presence of other points. If there exists no other point in the continuity of $m$ windows, then this query point on the path is classified as a line component, else it is classified as a potential region having symbols.

\item The parameter $t$ i.e., the height of orthogonal window is determined by traversing the path and finding the maximum distance from points present on contours, along the orthogonal direction to the point according to the following equation: 
\begin{equation}
\centering
  \small{t = MAX(\epsilon, \beta\ MAX(\forall p_{iN} dist(p_i, p_{iN})))}
\label{eq:find_t}
\end{equation}
where $p_{iN}$ are the points along the paths orthogonal to the query point $p_{i}$.
\item Visual elements of a P\&ID are largely connected (with few minor exceptions). In cases of small discontinuities in straight lines (pipes), the regions of discontinuity are marked as potential symbol regions. Even if the line terminates (at a symbol like a flange), and the slope changes by a significant amount, the terminal region of the line is marked as a potential symbol.
\end{itemize}

The individual regions obtained are plausible regions which can contain the symbols. Figure~\ref{fig:region_sampling} shows a small patch of a P\&ID illustrating the sampled points. Symbols are present over the straight lines. Junction points obtained over every path are used to segregate symbol regions, by removing the straight connecting lines.

\subsection{GCN based Symbol Classification}
\label{subsec:gcn-identification}
\noindent \textbf{Symbol Classes}: We use a set of symbol classes containing one clean sample image per class for symbol identification. We observe that these symbols exist in 4 different orientations each rotated by $90$ degrees. So, we augment the symbol images using rotation by $90$ degrees.

\begin{figure}[h]
\centering
\includegraphics[width=0.4\textwidth]{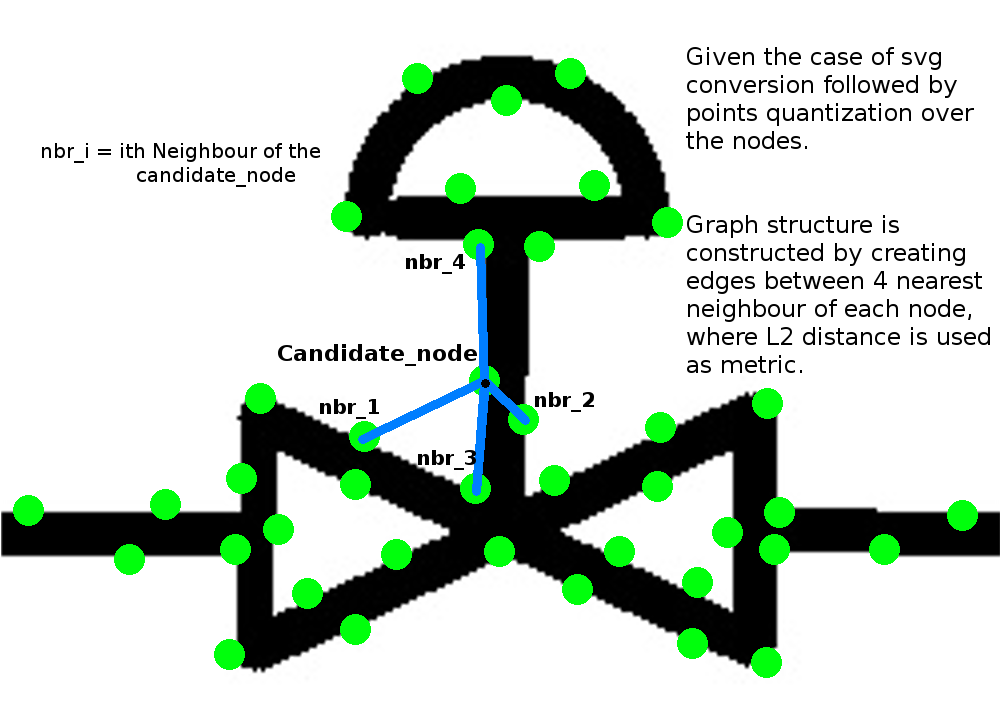}
\vspace{-3mm}
\caption{Figure demonstrating the structure of the graph formed by using $4$ nearest neighbours which are evaluated based on L2-distance over each point feature shown in green.}
\vspace{-3mm}
\label{fig:graph_explain}
\end{figure}

\begin{table*}[!ht]
\centering
\caption{Comparison of the performance of symbol classification using DGCNN and its variants having Arcface loss and Resnet-34 embeddings on the synthetic Dataset-P\&ID}
\label{tab:comp-symb-det}
\begin{tabular}{|c|c|c|c|c|c|c|c|c|c|}
\hline
\multirow{2}{*}{\textbf{Symbols}} & \multicolumn{3}{c|}{\textbf{DGCNN}} & \multicolumn{3}{c|}{\textbf{DGCNN + Arcface}} & \multicolumn{3}{c|}{\textbf{DGCNN + Resnet-34 + Arcface}} \\ \cline{2-10} 
 & \textbf{Precision} & \textbf{Recall} & \textbf{F1-score} & \textbf{Precision} & \textbf{Recall} & \textbf{F1-score} & \textbf{Precision} & \textbf{Recall} & \textbf{F1-score} \\ \hline
\textit{symbol1} & 0.7565 & 0.7005 & 0.7274 & 0.8669 & \textbf{0.8096} & \textbf{0.8373} & \textbf{0.8758} & 0.7960 & 0.8340 \\ \hline
\textit{symbol2} & 0.8161 & 0.8512 & 0.8333 & \textbf{0.9202} & \textbf{0.8975} & \textbf{0.9087} & 0.9005 & 0.8920 & 0.8962 \\ \hline
\textit{symbol3} & 0.7602 & 0.7238 & 0.7416 & 0.9119 & 0.7937 & 0.8487 & \textbf{0.9138} & \textbf{0.8043} & \textbf{0.8556} \\ \hline
\textit{symbol4} & 0.7360 & 0.7825 & 0.7585 & 0.8293 & \textbf{0.8708} & \textbf{0.8495} & \textbf{0.8441} & 0.8475 & 0.8458 \\ \hline
\textit{symbol5} & 0.8137 & 0.7630 & 0.7875 & 0.8809 & \textbf{0.9068} & 0.8937 & \textbf{0.8913} & 0.9022 & \textbf{0.8967} \\ \hline
\textit{symbol6} & 0.8316 & 0.7562 & 0.7921 & \textbf{0.9220} & \textbf{0.9292} & \textbf{0.9255} & 0.9146 & 0.9146 & 0.9146 \\ \hline
\textit{symbol7} & 0.7875 & 0.7478 & 0.7671 & \textbf{0.8938} & \textbf{0.8539} & \textbf{0.8734} & 0.8668 & 0.8310 & 0.8485 \\ \hline
\textit{symbol8} & 0.7520 & 0.8473 & 0.7968 & 0.7925 & 0.9031 & 0.8442 & \textbf{0.7947} & \textbf{0.9274} & \textbf{0.8560} \\ \hline
\textit{symbol9} & 0.6144 & 0.8366 & 0.7084 & 0.6672 & \textbf{0.9146} & \textbf{0.776} & \textbf{0.6714} & 0.8968 & 0.7679 \\ \hline
\textit{symbol10} & 0.8595 & 0.7355 & 0.7926 & \textbf{0.9296} & 0.9003 & 0.9147 & 0.9244 & \textbf{0.9178} & \textbf{0.9211} \\ \hline
\textit{symbol11} & 0.6786 & 0.8614 & 0.7591 & 0.7541 & \textbf{0.9219} & \textbf{0.8296} & \textbf{0.7583} & 0.9095 & 0.8271 \\ \hline
\textit{symbol12} & 0.7609 & 0.61 & 0.6771 & 0.8626 & 0.6691 & 0.7536 & \textbf{0.8692} & \textbf{0.6786} & \textbf{0.7622} \\ \hline
\textit{symbol13} & 0.8304 & 0.7907 & 0.8101 & 0.9087 & \textbf{0.8643} & \textbf{0.8860} & \textbf{0.9137} & 0.8412 & 0.8760 \\ \hline
\textit{symbol14} & 0.8601 & 0.8175 & 0.8382 & 0.8636 & \textbf{0.9076} & 0.8850 & \textbf{0.8687} & 0.9060 & \textbf{0.8869} \\ \hline
\textit{symbol15} & 0.7614 & 0.7537 & 0.7576 & 0.8660 & 0.9139 & 0.8893 & \textbf{0.8759} & \textbf{0.9348} & \textbf{0.9044} \\ \hline
\textit{symbol16} & 0.7802 & 0.8481 & 0.8128 & \textbf{0.9041} & 0.8452 & \textbf{0.8736} & 0.8771 & \textbf{0.8679} & 0.8724 \\ \hline
\textit{symbol17} & 0.6363 & 0.7968 & 0.7076 & \textbf{0.7176} & 0.8886 & \textbf{0.7940} & 0.7131 & \textbf{0.8904} & 0.7919 \\ \hline
\textit{symbol18} & 0.8159 & 0.8259 & 0.8209 & 0.8670 & 0.9288 & 0.8968 & \textbf{0.8784} & \textbf{0.9320} & \textbf{0.9044} \\ \hline
\textit{symbol19} & 0.7554 & 0.6301 & 0.6871 & \textbf{0.8617} & 0.6818 & \textbf{0.7612} & 0.8475 & \textbf{0.6858} & 0.7581 \\ \hline
\textit{symbol20} & 0.7844 & 0.7636 & 0.7739 & 0.8857 & \textbf{0.8645} & 0.8750 & \textbf{0.8985} & 0.8560 & \textbf{0.8767} \\ \hline
\textit{symbol21} & 0.8520 & 0.8193 & 0.8353 & 0.9335 & 0.9025 & 0.9177 & \textbf{0.9390} & \textbf{0.9044} & \textbf{0.9214} \\ \hline
\textit{symbol22} & 0.7872 & 0.8282 & 0.8072 & \textbf{0.8981} & 0.8899 & \textbf{0.8940} & 0.8702 & \textbf{0.9028} & 0.8862 \\ \hline
\textit{symbol23} & 0.7314 & 0.8093 & 0.7684 & 0.8152 & \textbf{0.8991} & 0.8552 & \textbf{0.8326} & 0.8938 & \textbf{0.8622} \\ \hline
\textit{symbol24} & 0.7196 & 0.7593 & 0.7389 & 0.8535 & 0.8325 & 0.8428 & \textbf{0.8596} & \textbf{0.8685} & \textbf{0.8641} \\ \hline
\textit{symbol25} & 0.7963 & 0.6293 & 0.7030 & 0.9140 & \textbf{0.6749} & \textbf{0.7765} & \textbf{0.9161} & 0.6702 & 0.7741 \\ \hline
\end{tabular}%

\end{table*}

\noindent\textbf{Dynamic Graph CNN for learning symbol representation}: The set of points generated along the periphery of obtained probable symbol regions are classified into one among the symbol classes using DGCNN. To illustrate, graph representation of filtered segregated regions is illustrated in Figure~\ref{fig:graph_explain}. Since every symbol lies on a line (pipe), we look at collinear line segments as defined in the previous section. Gaps or discontinuities between line segments associated with a single long line (pipe) are examined to check if there are symbols within the gaps. We use points inside these gap regions for the classification. 

Observing the number of line segments intersecting a particular gap region, can help in improving the robustness of classification as certain symbols only allow for certain connectivity patters (for example certain valves may always have connectivity to two lines corresponding to inlet and outlet ). However, to handle the cases where we do not get any connectivity information, all the independent contours orthogonal and close to the moving lines (pipes) are used for the classification.
We compute the following features for each point present on the contour:
\begin{itemize}
\item Coordinate information defining each point on contour.
\item Hu-moments~\cite{huang2010analysis} of each point are calculated and appended as point features. HU moments are a set of $7$ values which are invariant to image transformations like translation, scale, rotation and reflection. Since, the points are arranged sequentially, a window of $m$ (here, $6$) sequential points are used to calculate the seven HU moments for every feature point. 
\begin{equation}
  h_0 = \eta_{20} + \eta_{02}
  \label{eqn:hu1}
\end{equation}
\begin{equation}
  h_1 = (\eta_{20} - \eta_{02})^{2} + 4\eta_{11}^{2}
\label{eqn:hu2}
\end{equation}
\begin{equation}
  h_2 = (\eta_{30} - 3\eta_{12})^{2} + (3\eta_{21} - \eta_{03})^{2}
 \label{eqn:hu3}
\end{equation}
\begin{equation}
  h_3 = (\eta_{30} + \eta_{12})^{2} + (\eta_{21} + \eta_{03})^{2}
\label{eqn:hu4}
\end{equation}
\begin{equation}
\begin{split}
  h_4 = (\eta_{30} - 3\eta_{12})(\eta_{30} + \eta_{12}) [(\eta_{30} + \eta_{12})^{2}-\\
  3(\eta_{21} + \eta_{03})^{2}]+ \\
  (3\eta_{21} - \eta_{03})(\eta_{21} + \eta_{03}) [3(\eta_{30} + \eta_{12})^2 -\\
   (\eta_{21} + \eta_{03})^2]
\end{split}
\label{eqn:hu5}
\end{equation}
\begin{equation}
\begin{split}
  h_5 = (\eta_{20} - \eta_{02}) [(\eta_{30} + \eta_{12})^2-(\eta_{21} + \eta_{03})^2] + \\
  4\eta_{11}(\eta_{30} + \eta_{12})(\eta_{21} + \eta_{03})
\end{split}
\label{eqn:hu6}
\end{equation}
\begin{equation}
\begin{split}
  h_6 = (3\eta_{21}-\eta_{03})(\eta_{30} + \eta_{12})[(\eta_{30} + \eta_{12})^2-\\
  3(\eta_{21} + \eta_{03})^2]- \\
  (\eta_{30} -3 \eta_{12})(\eta_{21} + \eta_{03})[3(\eta_{30} + \eta_{12})^2-\\
  (\eta_{21} + \eta_{03})^2]
\end{split}
\label{eqn:hu7}
\end{equation}
Equations.~\ref{eqn:hu1},~\ref{eqn:hu2},~\ref{eqn:hu3},~\ref{eqn:hu4},~\ref{eqn:hu5},~\ref{eqn:hu6} and ~\ref{eqn:hu7} represent the equations for the calculation of seven HU moments. The $\eta_{ij}$ are the normalized central moments, which are computed as follows:
\begin{equation}
  \eta_{ij} = \frac{\mu_{ij}}{\mu_{00}^{(i+j)/2+1}}
  \label{eqn:eta}
\end{equation}
\begin{equation}
  \mu_{ij} = \Sigma_x\Sigma_y(x-\Bar{x})^i(y-\Bar{y})^j
  \label{eqn:mu}
\end{equation}
\item DGCNN authors~\cite{dgcnn} claim that with fewer than $512$ points in the graph, the performance of the network deteriorates dramatically. Therefore, we have fixed the number of points to be $1024$ for our case. To maintain an equal number of data points for every symbol, we interpolate points over different loops based on their length such that the total number of points are constant.
\end{itemize}

By appending the seven HU-moment features to three coordinate value features, we get 9 features per point. These points are used to train the DGCNN network for classification. Figure~\ref{fig:network_arch} shows the architecture of the network used for symbol classification. The input to network, in our case, is 1024 points with each point having 9 features $(F =9)$. This input is passed through three sequential edge-convolution layers. The outputs of all these three edge convolution layers are appended globally to form a 1-D global descriptor, which is used to generate classification scores for $C$ classes. The embeddings obtained at the final trained global layer are stored and used for comparison at inference.

\noindent\textbf{Appending Resnet-34 with Dynamic Graph CNN}:
To make symbol classification more robust, we append global embeddings of images from the ResNet-34 network. To achieve this, the ResNet-34 model is trained over the images of the graph of sampled points by initializing the network with imagenet pre-trained weights. The modified network architecture, as shown in Figure~\ref{fig:dgcnn_resnet}, uses a similar methodology as used in DGCNN, however at the global embedding, the image visual features are additionally appended.

\subsection{Learning and Inference}
\label{subsec:inference}
\noindent \textbf{Loss Function}: The cross-entropy loss paired with a softmax layer is the standard architecture used for most real world classification tasks which works by creating separation between embeddings of different classes. However, in case of symbols which are visually very similar to other classes, this loss causes the network to mis-classify if there occurs even a slight deviation in the test sample embeddings. We address this by utilizing Arcface loss~\cite{deng2019arcface} which trains the network to generate class embeddings with an additional angular margin between them and increases the discriminating power of the classifier. Equation.~(\ref{arcfaceloss}) represents the Arcface loss for $C$ classes with additional $m$ margin. The loss function differs for each batch of $N$ examples, where $y_{i}$ is the  ground truth for the $i^{th}$ sample in the batch.

\vspace{-2mm}
\begin{equation}
\label{arcfaceloss}
L_{Arcface} = -\frac{1}{N}\sum_{i=1}^{N}log\frac{e^{s(cos(\theta_{y_{i}}+m))}}{e^{s(cos(\theta_{y_{i}}+m))} + \sum_{j=1, j{\neq}y_{i}}^{C}{e^{s (cos (\theta_{j}))}}}
\end{equation}

\noindent \textbf{Augmentation Policy}: Data augmentation is routinely employed to reduce over-fitting and obtain better generalization performance. However, while considering symbols having very minute inter-class variations, the standard augmentation techniques do not prove very beneficial. One underlying reason for this is that the most augmentation techniques perform uniform changes in the resulting augmented data. This can be addressed by incorporating different modifications for different sub-parts of each example image. In our approach, the augmentations are generated for each symbol image by incorporating affine transformations to each sub-part of the image with rotation ranging from angle $-20$\degree to $20$\degree, scaling parameter ranging from $0.9$ to $1.1$ and shear parameter ranging from $-0.05$ to $0.05$. The resultant transformation matrix is used for transformation over a window of $m$ sequential points. The transformation matrices over the sequential windows have parameters in the range $-0.01$ to $0.01)$. This augmentation policy gives more flexibility to the network to adapt for local changes in addition to retaining the advantages of the traditional augmentation approach.

\noindent \textbf{Inference}:
To increase the robustness of the model at inference time, we take each symbol region image, cropped from the P\&ID sheet, at two scales with $4$ different orientations. The two scales are chosen in a way that each symbol region image has a minimum width of $300$ and $600$ px respectively. Hence, for each given symbol-region image, we produce $8$ different embeddings which are compared using the cosine distance against the symbol embedding directory for the entire set of symbols. The final class label for the given symbol-region image is assigned based upon  majority voting.

\section{Experimental Results and Discussions}
\label{sec:experimental-results}

\subsection{Data}
\begin{figure}[!ht]
\centering
\includegraphics[width=0.35\textwidth]{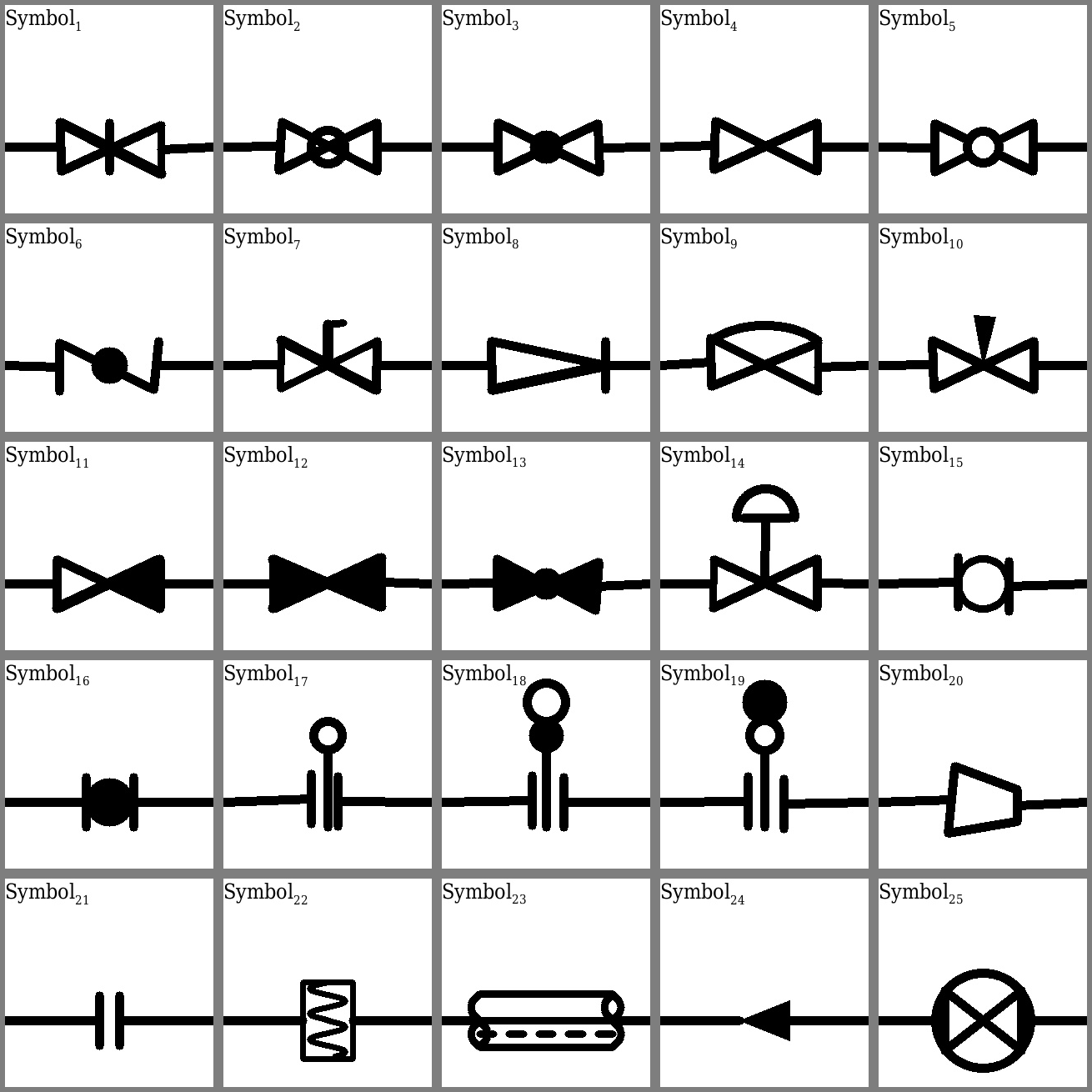}
\caption{List of symbols ($Symbol_{1}$ - $Symbol_{25}$) used for training and evaluation of our proposed method.}
\label{fig:symbol_list}
\end{figure}

We evaluate the performance of our proposed one-shot symbol recognition method on a synthetic dataset called Dataset-P\&ID which consists of $100$ P\&ID sheets in the test-set. These P\&ID sheets contain various components (symbols) attached to a network of pipelines. In this paper, we aim to localize and subsequently classify a set of $25$ symbols, as shown in Figure~\ref{fig:symbol_list}, from these P\&ID sheets. We also compare our proposed method against the method by Rahul et. al~\cite{auto_pid2} on a private dataset of $12$ real P\&ID sheets for symbol recognition.

\begin{table}[!h]
\centering
\caption{Performance of Symbol-Region Segmentation}
\begin{tabular}
{|M{2cm}|M{1.3cm}|M{1.3cm}|M{1.3cm}|M{1.3cm}|}
 \hline
\textbf{Total Symbols Regions} & \textbf{Correct Regions Localized} & \textbf{False Regions Localized}  & \textbf{Missing Regions} \\
\hline
 10630 & 10482 & 43 & 148\\
\hline 
\end{tabular}
\label{tab:symb-region}
\vspace{-3mm}
\end{table}
\vspace{-3mm}

\begin{table}
\centering
\caption{Comparison of proposed OSSR-PID which performs one-shot learning against Rahul et al~\cite{auto_pid2} which uses a large training number of training images. Results are on a real world P\&ID dataset}
\label{table:tableComp}
\begin{tabular}{|p{1.25cm}|p{0.75cm}|p{0.75cm}|p{0.75cm}|p{0.75cm}|p{0.75cm}|p{0.75cm}|}
\hline & \multicolumn{2}{c|}{\textbf{Precision}}   & \multicolumn{2}{c|}{\textbf{Recall}} & \multicolumn{2}{c|}{\textbf{F1-score}} \\ 
\hline \textbf{Symbols} & \multicolumn{1}{c|}{\textbf{\cite{auto_pid2} }} & \multicolumn{1}{c|}{\textbf{Ours}} & \multicolumn{1}{c|}{\textbf{\cite{auto_pid2} }} & \multicolumn{1}{c|}{\textbf{Ours}} & \multicolumn{1}{c|}{\textbf{\cite{auto_pid2} }} & \multicolumn{1}{c|}{\textbf{Ours}}  \\ 
\hline

\hline
\textbf{Bl-V} & \textbf{0.925} & 0.913 & \textbf{0.936} & 0.896 & \textbf{0.931} & 0.904 \\
\textbf{Ck-V} & \textbf{0.941} & 0.911 & \textbf{0.969} & 0.902 & \textbf{0.955} & 0.906\\
\textbf{Ch-sl} & \textbf{1.000} & 0.99 & 0.893 & \textbf{0.902} & 0.944 & 0.944\\
\textbf{Cr-V} & 1.000 & 1.000 & \textbf{0.989} & 0.98 & \textbf{0.995} & 0.990\\
\textbf{Con} & \textbf{1.000} & 0.875 & 0.905 & \textbf{0.922} & \textbf{0.950} & 0.897\\
\textbf{F-Con} & \textbf{0.976} & 0.862 & 0.837 & \textbf{0.947} & 0.901 & \textbf{0.903}\\
\textbf{Gt-V-nc} & 0.766 & \textbf{0.894} & 1.000 & 1.000 & 0.867 & \textbf{0.944}\\
\textbf{Gb-V} & \textbf{0.888} & 0.871 & \textbf{0.941} & 0.884 & \textbf{0.914} & 0.877\\
\textbf{Ins} & 1.000 & 1.000 & \textbf{0.985} & 0.982 & \textbf{0.992} & 0.990\\
\textbf{GB-V-nc} & 1.000 & 1.000 & 0.929 & \textbf{0.977} & 0.963 & \textbf{0.988}\\
\textbf{Others} & \textbf{0.955} & 0.927 & \textbf{1.000} & 0.942 & \textbf{0.970} & 0.934\\
\hline

\end{tabular}
\label{tab:comparison}
\vspace{-4mm}
\end{table}

\subsection{Results}
First, we present the results of symbol-region localization in 
Table\ref{tab:symb-region}. Here, the emphasis is more on high recall to make sure that no symbol-region is missed prior to its classification. The results demonstrate that the proposed method for symbol-region segmentation performs remarkably well with only few misses ($<$0.5\% approx.). Next, we evaluate our one-shot symbol recognition method using DGCNN (with and without Arcface loss and Resnet-34) on the synthetic P\&ID sheets and tabulate results in Table~\ref{tab:comp-symb-det}. As it is evident from Table~\ref{tab:comp-symb-det}, our proposed OSSR-PID method is able to recognize different symbols using only one prototypical example image per symbol class with excellent precision, recall and F1-score values. We also computed average accuracy of symbol classification for these experiments and obtained $77.07\%$ accuracy of symbol classification using DGCNN with categorical cross-entropy loss. However, we observed significant improvement in accuracy using Arcface loss where the average accuracy of symbol classification obtained is $85.78\%$, and while using DGCNN+ResNet-34+Arcface loss we get comparable accuracy ($85.98\%$) with very slight improvement. Please note that we use $1024$ points for the classification using DGCNN and the number  (\begin{math}k\end{math}) of nearest neighbors for EdgeConv block is set to be $20$. 

\begin{figure}[h]
\centering
\includegraphics[width=0.4\textwidth]{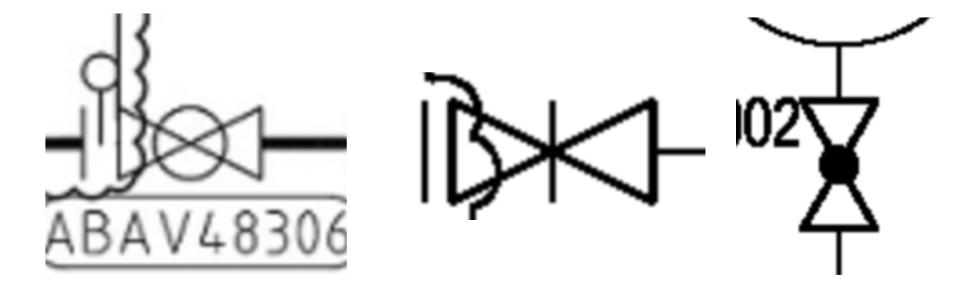}
\caption{Figure showing failure cases for symbol recognition by our proposed method.
}
\label{fig:fail_case}
\end{figure}

Further, we compare OSSR-PID against prior-work~\cite{auto_pid2} and present the results on real P\&ID sheets dataset in Table~\ref{tab:comparison}. For this comparison, we use the same set of symbols as used by Rahul et. al~\cite{auto_pid2}. The single instance example of each class is taken from real P\&ID sheet, and all the factors are fixed as described in ~\cite{auto_pid2}. As it is evident from Table~\ref{tab:comparison} that OSSR-PID performs remarkably well and comparable to the earlier method~\cite{auto_pid2}. However, it should be noted that OSSR-PID requires only one single training image per symbol class and hence, and offers comparable performance to ~\cite{auto_pid2} which is fully supervised and needs a large amount of annotated training data for each symbol class. At last, we also show some failure cases of our proposed method where it mis-classifies the symbols due to the noise and clutter present around the symbol of interest.

\section{Conclusion}
\label{sec:conclusion}
In this paper, we present a technique for one-shot localization and recognition of symbols in a P\&ID sheet by using a path sampling technique for extraction of probable symbol regions and a DGCNN for symbol classification. The technique is specialized for line-drawn engineering drawings and requires just one instance of a symbol image to yield impressive test recognition accuracy. The technique was tested on $100$ synthetic P\&ID diagrams depicting oil and gas flow schematics, which will be released for future research. The current one-shot learning model offers comparable performance to prior fully supervised techniques, however, it still needs to be trained on inclusion of new class symbols. This forms the basis of future followup work on this topic, in which re-training can be fully avoided, i.e. employ zero-shot learning, to be used over different seen and unseen symbol classes.
\bibliographystyle{IEEEtran}
\bibliography{IEEEabrv,mybibfile}

\end{document}